\documentclass[a4paper]{article}

\usepackage{INTERSPEECH2015}

\usepackage{graphicx,hyperref}
\usepackage{amssymb,amsmath,bm,color}
\usepackage{textcomp}

\hypersetup{pdftoolbar=true,pdfpagemode=UseNone,pdfstartview=FitH,
  colorlinks=true,extension=}

\newcounter{multifig}

\newcommand{\figcaption}[1]% #1 = text
{\stepcounter{multifig}
\addcontentsline{lof}{figure}{\string\numberline {\arabic{multifig}}{\ignorespaces #1}}
Figure \arabic{multifig}: #1}

\ninept

\title{End-to-End Text-Dependent Speaker Verification}

\twoauthors{Georg Heigold$^*$\thanks{$^*$ Work done while the author was at Google.}}{DFKI, Germany\\
  {\small \tt georg.heigold@dfki.de}
 }
  {Ignacio Moreno, Samy Bengio, Noam Shazeer}{Google Inc., USA \\
  {\small \tt \{elnota,bengio,noam\}@google.com} 
 }

\begin{document}

  \maketitle
  
   \setlength{\belowdisplayskip}{7.5pt} \setlength{\belowdisplayshortskip}{7.5pt}
\setlength{\abovedisplayskip}{7.5pt} \setlength{\abovedisplayshortskip}{7.5pt}
  
    \begin{abstract}  
  % What?
  In this paper we present a data-driven, integrated approach to speaker 
  verification, which maps a test utterance and a few reference utterances
  directly to a single score for 
  verification and jointly optimizes the system's components
  using the same evaluation protocol and metric as at test time.
  % Why?
  Such an approach will result in simple and efficient systems, requiring
  little domain-specific knowledge and making few model assumptions.
  % How?
  We implement the idea by formulating the problem as a
  single neural network architecture, including the estimation of a speaker 
  model on only a few utterances, and evaluate it on 
  our internal "Ok Google" benchmark for text-dependent speaker verification.
  % High-level summary.
  The proposed approach appears to be very effective for big 
  data applications like ours that require highly accurate, easy-to-maintain 
  systems with a small footprint.
  \end{abstract}
  \noindent{\bf Index Terms}: speaker verification, end-to-end training, deep learning.

    \section{Introduction}

  % what is speaker verification
  Speaker verification is the process of verifying, based on 
  a speaker's known utterances, whether an utterance belongs to
  the speaker. When the lexicon of the spoken utterances is 
  constrained to a single word or phrase across all users, the process
  is referred to as global password text-dependent speaker verification. 
  By constraining the lexicon, text-dependent speaker verification aims to
  compensate for phonetic variability, which poses a significant challenge in speaker 
  verification~\cite{Aronowitz:2011}. 
  At Google, we are interested in text-dependent speaker verification with the
  global password "Ok Google." 
  The choice of this particularly short, approximately 0.6 seconds long global 
  password relates to the Google Keyword Spotting system~\cite{Prabhavalkar:2015} 
  and Google VoiceSearch~\cite{Schalkwyk:2010} and facilitates the combination of the systems.
  
  In this paper, we propose to directly map a test utterance together with a few utterances
  to build the speaker model, to a single score for verification. All the components
  are jointly optimized using a verification-based loss following the standard speaker
  verification protocol. 
  Compared to existing approaches, such an end-to-end approach may have several advantages,
  including the \emph{direct modeling} from utterances, which allows
  for capturing long-range context and reduces the complexity
  (one vs. number of frames evaluations per utterance),
  and the \emph{direct and joint estimation}, which can lead to better and 
  more compact models.
  Moreover, this approach often results in considerably simplified systems requiring fewer 
  concepts and heuristics.
  
  More specifically, the contributions of this paper include:\vspace{-1ex}
  \begin{itemize}\setlength\itemsep{-0.2em}
    \item formulation of end-to-end speaker verification architecture, including the 
    estimation of a speaker model on a few utterances (Section~\ref{sec:e2e});
    \item empirical evaluation of end-to-end speaker verification, including
    comparison of frame (i-vectors, d-vectors) and utterance-level representations 
    (Section~\ref{sec:frame-vs-utterance}) and analysis of the end-to-end loss
    (Section~\ref{sec:losses});
    \item empirical comparison of feedforward and recurrent neural networks 
    (Section~\ref{sec:architectures}).
  \end{itemize}\vspace{-1ex}
  This paper focuses on text-dependent speaker verification for small footprint systems,
  as discussed in~\cite{Chen:2015}.
  But the approach is more general and could be used similarly for text-independent 
  speaker verification.
  
  % state of the art
In previous studies, the verification problem is broken down
into more tractable, but loosely connected subproblems. For
example, the combination of i-vector and probabilistic linear
discriminant analysis (PLDA)~\cite{Kenny:2010,Dehak:2011} 
has become the dominant approach, both for text-independent speaker 
verification~\cite{Reynolds:2000,Kenny:2007,Kenny:2010,Dehak:2011}
and text-dependent speaker verification~\cite{Aronowitz:2012,Stafylakis:2013,Larcher:2013}. 
Hybrid approaches that include deep learning based components
have also proved to be beneficial for text-independent speaker
recognition~\cite{Garcia:2014,Lei:2014,Richardson:2015}. For small footprint systems, however,
a more direct deep learning modeling may be an attractive 
alternative~\cite{Variani:2014,Chen:2015}. 
To the best of our knowledge, recurrent neural
networks have been applied to related problems such as speaker
identification~\cite{Parveen:2000} and language identification~\cite{Gonzalez:2014}, 
but not to the speaker verification task.
The proposed neural network architecture can be thought of as joint optimization
of a generative-discriminative hybrid and is in the same spirit as deep 
unfolding~\cite{Hershey:2014} for adaptation.

  The remainder of the paper is organized as follows.
  Section~\ref{sec:speaker-verification} provides a brief overview 
  of speaker verification in general.
  Section~\ref{sec:d-vector} describes the d-vector approach.
  Section~\ref{sec:e2e} introduces the proposed end-to-end approach
  to speaker verification.
  An experimental evaluation and analysis can be found in Section~\ref{sec:experiments}.
  The paper is concluded in Section~\ref{sec:summary}.

    \section{Speaker Verification Protocol}\label{sec:speaker-verification}

  The standard verification protocol can be divided into the three steps:
  training, enrollment, and evaluation, which we describe in more detail next.

 \vspace{-3ex}\paragraph*{Training}
  In the training stage, we find a suitable internal speaker representation from the utterance,
  allowing for a simple scoring function.
  In general, this representation depends on
  the type of the model (e.g., Gaussian subspace model or deep neural network),
  the representation level (e.g., frame or utterance), and
  the model training loss (e.g., maximum likelihood or softmax).
  State-of-the art representations are a summary of frame-level information, such as
  i-vectors~\cite{Reynolds:2000,Kenny:2007} and d-vectors (Section~\ref{sec:d-vector}).
 
 \vspace{-3ex}\paragraph*{Enrollment}
  In the enrollment stage, a speaker provides a few utterances (see Table~\ref{tab:datasets}),
  which are used to estimate a speaker model.
  A common choice is to average the i-vectors~\cite{Greenberg:2014} 
  or d-vectors~\cite{Variani:2014,Chen:2015} of these utterances.
  
   \vspace{-3ex}\paragraph*{Evaluation}
  During the evaluation stage, the verification task is performed and the system is evaluated.
  For verification, the value of a scoring function of the utterance $X$ and the test speaker $spk$, 
  $S(X,spk)$, is compared against a pre-defined threshold. 
  We accept if the score exceeds the threshold, i.e., the utterance $X$ comes from
  speaker $spk$, and reject otherwise. 
  In this setup, two types of error can occur: false reject and false accept.
  Clearly, the false reject rate and the false accept rate depend on the threshold.
  When the two rates are equal, the common value is called equal error rate (EER).

  A simple scoring function is the cosine similarity between
  the speaker representation $f(X)$ of an evaluation utterance $X$ (see paragraph "Training") 
  and the speaker model $m_{spk}$ (see paragraph "Enrollment"):
  \vspace{-1ex}\begin{equation*}\vspace{-1ex}
    S(X,spk) =[f(X)^{\top}m_{spk}] / [\|f(X)\|\phantom{\cdot}\|m_{spk}\|].
  \end{equation*}
  PLDA has been proposed as a more refined, data-driven scoring approach.

    \section{D-Vector Baseline Approach}\label{sec:d-vector}
  
\begin{figure}[b]
\vspace{-4ex}
    \centering
      \includegraphics[width=0.55\linewidth]{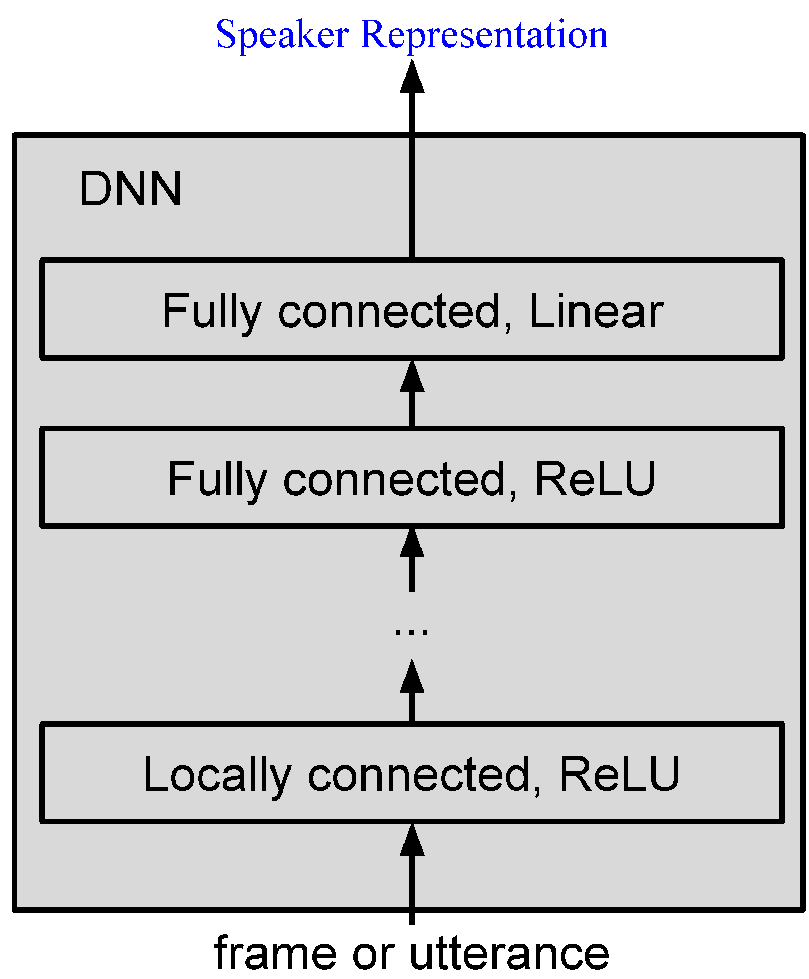}
           \vspace{-2ex}
     \caption{\label{fig:dnn} \it Deep neural network (DNN) with a locally-connected layer 
     followed by fully-connected layers.}\vspace{0ex}
  \end{figure}

  D-vectors are derived from a deep neural network (DNN), as speaker
  representation of an utterance.
  A DNN consists of the successive application of several non-linear functions 
  in order to transform the speaker utterance into a vector where a decision 
  can be easily made.
  Fig.~\ref{fig:dnn} depicts the topology of our baseline DNN.
  It includes a locally-connected layer~\cite{Chen:2015} and several fully connected layers.
  All layers use ReLU activation except the last, which is linear.
  During the training stage, the parameters of the DNN are optimized using the softmax loss, which, for convenience,
  we define to comprise
  a linear transformation with weight vectors $w_{spk}$ and biases $b_{spk}$, followed
  by the softmax function and the cross-entropy loss:
  \vspace{-1ex}\begin{equation*}\vspace{-1ex}
    l_{\text{softmax}} = -\log \frac{\exp(w_{spk}^{\top}y+b_{spk})}{\sum_{\tilde{spk}}\exp(w_{\tilde{spk}}^{\top}y+b_{\tilde{spk}})}
\end{equation*}
where the activation vector of the last hidden layer is denoted by $y$ and
$spk$ denotes the correct speaker.
 The normalization is over all competing training speakers $\tilde{spk}$.
  
After the training stage is completed, the parameters of the DNN are fixed. 
Utterance d-vectors are obtained by averaging
the activation vectors of the last hidden layer for all frames of
an utterance. Each utterance generates one d-vector. For
enrollment, the speaker model is given by the average over
the d-vectors of the enrollment utterances. Finally, during the
evaluation stage, the scoring function is the cosine similarity
between the speaker model d-vector and the d-vector of a test
utterance.  
 
 Criticism about this baseline approach includes the limited context of 
 the d-vectors derived from (a window of) frames and the type of the loss.
 The softmax loss attempts to discriminate between the true speaker and
 all competing speakers but does not follow the standard
 verification protocol in Section~\ref{sec:speaker-verification}.
As a result, heuristics and scoring normalization techniques becomes necessary to
compensate for inconsistencies.
 Moreover, the softmax loss does not scale well with more data
 as the computational complexity is linear in the number of training speakers
 and requires a minimum amount of data per speaker to estimate the speaker-specific
 weights and biases. 
 The complexity issue (but not the estimation issue) 
 can be alleviated by candidate sampling~\cite{Jean:2014}. 
 
 Similar concerns can be expressed over the alternative
speaker verification approaches, where some of the component blocks are either loosely connected or not directly optimized following the speaker verification protocol. 
For example,
GMM-UBM~\cite{Reynolds:2000} or i-vector models does not directly optimize a
verification problem; the PLDA~\cite{Kenny:2010} model is not followed a re-
finement of the i-vector extraction; or long contextual features
may be ignored by frame-based GMM-UBM models~\cite{Reynolds:2000}.

    \section{End-To-End Speaker Verification}\label{sec:e2e}
\begin{figure}[t]\vspace{-0ex}
    \centering
      \includegraphics[width=0.74\linewidth]{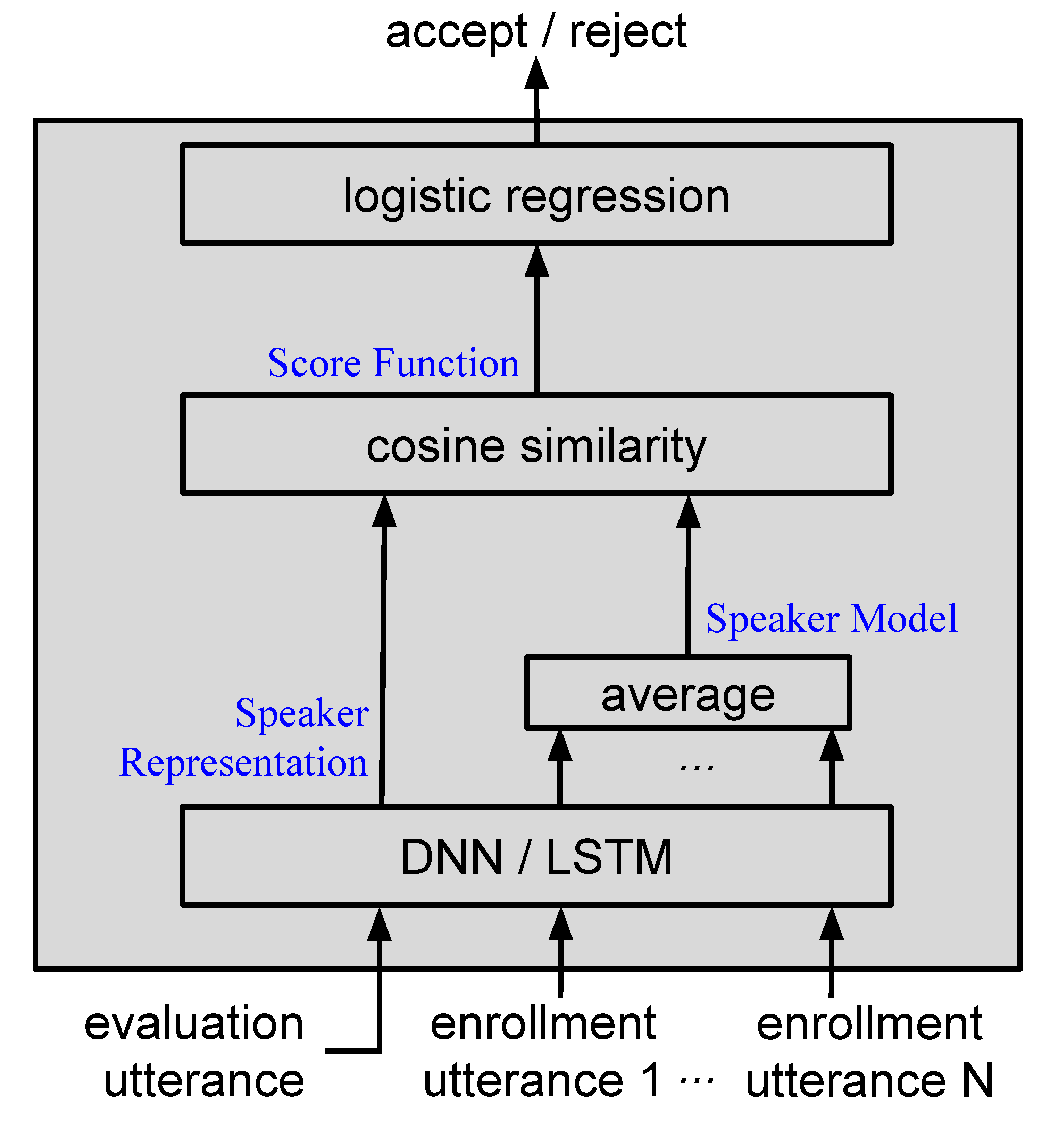}
      \vspace{-2ex}
      \caption{\label{fig:end-to-end} \it End-to-end architecture, combining training of 
      speaker representations (DNN/LSTM box), 
      estimation of a speaker model based on up to $N$ "enrollment" utterances (in blue),
      and verification (cosine similarity/logistic regression boxes).}
      \vspace{-2ex}
  \end{figure}

  In this section, we formulate the speaker verification problem as a single
  network architecture. Following Section~\ref{sec:speaker-verification},
  this architecture consists of a training component to compute the speaker representations, 
  an enrollment component to estimate the speaker model, and 
  an evaluation component with a suitable loss function for optimization,
  see Fig.~\ref{fig:end-to-end}.
  
\begin{figure}[t]
  \vspace{-0ex}
    \centering
     \includegraphics[width=0.66\linewidth]{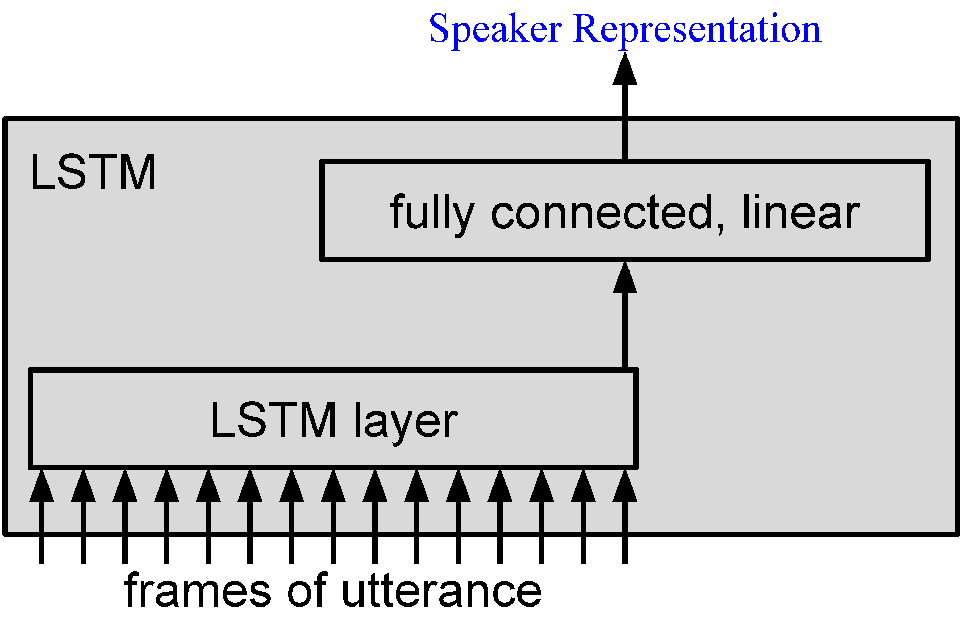}
     \vspace{-2ex}
     \caption{\label{fig:lstm} \it Long short-term memory recurrent neural network (LSTM) with a single output.}\vspace{-2ex}
  \end{figure}

  % embedding model
  We use neural networks to obtain the speaker representation of an utterance.
  The two types of networks we use in this work are 
  depicted in Figs.~\ref{fig:dnn} and~\ref{fig:lstm}:
  a deep neural network (DNN) with locally-connected and fully connected layers 
  as our baseline DNN in Section~\ref{sec:d-vector} and 
  a long short-term memory recurrent neural network (LSTM)~\cite{Hochreiter:1997,Sak:2014} 
  with a single output.
  DNNs assume a fixed-length input. To comply with this constraint, 
  we stack the frames of a sufficiently large window of fixed length over the utterance
 and use them as the input. This trick is not 
  needed for LSTMs but we use the same window of frames for better comparability. 
  Unlike vanilla LSTMs which have multiple outputs, we only connect the last output
  to the loss to obtain a single, utterance-level speaker representation.
  
  % speaker model
  The speaker model is the average over a small number of "enrollment" representations
  (Section~\ref{sec:speaker-verification}). We use the same network to compute
  the internal representations of the "test" utterance and of the utterances for the speaker model.
  The actual number of utterances per speaker available in training typically is much larger
  (a few hundred or more) than in enrollment (fewer than ten), see Table~\ref{tab:datasets}.
  To avoid a mismatch, we sample for each training utterance 
  only a few utterances from the same speaker to build the speaker model at training time.
  In general, we cannot assume to have $N$ utterances per speaker. To allow
  for a variable number of utterances, we pass a weight along with the utterance
  to indicate whether to use the utterance.
  
  % loss
  Finally, we compute the cosine similarity between the speaker representation and the 
  speaker model, $S(X,spk)$, and feed 
  it to a logistic regression including a 
  linear layer with a bias.
  The architecture is optimized using the end-to-end loss
  \vspace{-1ex}\begin{equation}\label{eq:e2e-loss}\vspace{-1ex}
    l_{e2e} = -\log p(target)
  \end{equation}
  with the binary variable $target\in\{\text{accept},\text{reject}\}$,
  $p(\text{accept})=(1+\exp(-w S(X,spk)-b))^{-1}$, and $p(\text{reject})=1-p(\text{accept})$.
  The value $-b/w$ corresponds with the verification threshold.
  
%  % similar architectures
%  This architecture combines a discriminative and a generative model, for speaker 
%  verification and the speaker model, respectively, and
%  is  similar to~\cite{Mairal:2012} or deep unfolding~\cite{Hershey:2014}.
%  Here, however, no unfolding is required because an analytic solution for the
%  generative speaker model exists.
  
  % input
  The input of the end-to-end architecture are $1+N$ utterances, i.e., an
  utterance to be tested and up to $N$ different utterances of the same speaker
  to estimate the speaker model.
  To achieve a good tradeoff between data shuffling and memory,
  the input layer maintains a pool of utterances to sample $1+N$ utterances from
  for each training step and gets refreshed frequently for better data shuffling.
%  Fig.~\ref{fig:end-to-end-input} illustrates the utterance sampling in the input layer.
%  \begin{figure}
%    \centering
%    \includegraphics[width=0.48\textwidth]{figures/End-To-End_Speaker_Verification_Input}
%    \caption{\it Illustration of the sampling in the end-to-end input layer.}
%    \label{fig:end-to-end-input}
%  \end{figure}
  As a certain number of utterances of the same speaker is needed for the speaker model,
  the data is presented in small groups of utterances of the same speaker.
%  Furthermore, the samples of a batch all come from different pools to obtain better
%  data shuffling, as depicted in Fig.~\ref{fig:shuffling}.
%  \begin{figure}
%    \centering
%    \includegraphics[width=0.35\textwidth]{figures/Shuffling}
%    \caption{\it Illustration of shuffling within a batch across utterances.}
%    \label{fig:shuffling}
%  \end{figure}
%  Last but not least, asynchronous SGD~\cite{Dean:2012} with many workers
%  also adds a lot of stochasticity.
%  We started with pre-generated tuples which was clearly suboptimal.
%  However, we have not evaluated how much each of the shuffling techniques
%  above actually contributes.
  
The end-to-end architecture allows for direct optimization of the evaluation metric using the standard evaluation protocol
  with consistent speaker models. 
  Conceivably, this will result in better accuracy without the need for 
  heuristics and post-processing steps, for example, score normalization.
  Moreover, the approach scales well as it neither depends on the number of training 
  speakers and nor requires a minimum number of utterances per speaker.
  
    \section{Experimental Evaluation}\label{sec:experiments}
  We evaluate the proposed end-to-end approach on our internal "Ok Google" benchmark.

%\vspace{-1ex}
  \subsection{Data Sets \& Basic Setup}\label{sec:data+setup}
%\vspace{-1ex}
We tested the proposed end-to-end approach on a set ?Ok Google? utterances collected from anonymized voice search logs.
  For improved noise robustness, we perform multistyle training.
  The data were augmented by artificially adding in car and cafeteria noise at various SNRs, 
  and simulating different distances between the speaker and the microphone,
  see~\cite{Prabhavalkar:2015} for further details.
  Enrollment and evaluation data include only real data. 
  Table~\ref{tab:datasets} shows some data set statistics.
  \begin{table}[h]\vspace{-2ex}
    \centering
    \caption{\label{tab:datasets} Data set statistics.}
    \begin{tabular}{|l|r|r|r|}
      \hline
                      & \#utterances & \#speakers & \#utts / spk \\
                      & (\#augmented) & & \\
      \hline\hline
      train\_2M & 2M (9M)         &  4k     & $>$500   \\
      train\_22M & 22M (73M)              & 80k   & $>$150     \\
      \hline
      enrollment      & 18k               &  3k     & 1-9   \\
      evaluation      & 20k               &  3k      & 3-5 \\
      \hline
%      enrollment, noisy      & 18k  / 72k              &  3k     & 2-57   \\
%      evaluation, noisy      & 20k  / 82k              &  3k      & 3-5? \\
%      \hline
    \end{tabular}
    \vspace{-1ex}
  \end{table}
  
  The utterances are forced aligned to obtain the "Ok Google" snippets.
  The average length of these snippets is around 80 frames, for a frame rate of 100 Hz.
  Based on this observation, we extracted the last 80 frames from each snippet,
  possibly padding or truncating frames at the beginning of the snippet.
  The frames consist of 40 log-filterbanks (with some basic spectral subtraction) each.

  For DNNs, we concatenate the 80 input frames, resulting in a 80x40-dimensional
  feature vector. Unless specified otherwise, the DNN consists of 4 hidden layers. 
  All hidden layers in the DNN have 504 nodes and use ReLU
  activation except the last, which is linear.
  The patch size for the locally-connected layer of the DNN is 10$\times$10.  
  For LSTMs, we feed the 40-dimensional feature vectors frame by frame.
  We use a single LSTM layer with 504 nodes without a projection layer.
  The batch size is 32 for all experiments.

  Results are reported in terms of equal error rate (EER), without and with 
  t-norm score normalization~\cite{Auckenthaler:2000}.

%\vspace{-1ex}
  \subsection{Frame-Level vs. Utterance-Level Representation}\label{sec:frame-vs-utterance}
%\vspace{-1ex}
  First, we compare frame-level and utterance-level speaker representations,
  see Table~\ref{tab:clean-frame-utterance}.
  Here, we use a DNN as described in Fig.~\ref{fig:dnn} with
  a softmax layer and trained on train\_2M (Table~\ref{tab:datasets}) with
  50\% dropout~\cite{Hinton:2012b} in the linear layer.
  The utterance-level approach outperforms the frame-level approach by 30\%.
  Score normalization gives a substantial performance boost (up to 20\% relative) 
  in either case.
    \begin{table}[h]\vspace{-2ex}
    \centering
    \caption{\label{tab:clean-frame-utterance} Equal error rates for
    frame-level and utterance-level speaker representations.}
    \begin{tabular}{|l|l|r|r|}
      \hline
              &   & \multicolumn{2}{c|}{EER (\%)} \\
      level       & system & raw & t-norm \\
      \hline\hline
      frame & i-vector~\cite{Dehak:2011} & 5.77 & 5.11 \\
                         & i-vector+PLDA~\cite{Garcia:2011} & 4.66 & 4.89 \\
		        & DNN, softmax~\cite{Chen:2015} & 3.86 & 3.32 \\
      utterance & DNN, softmax & 2.90 & 2.28 \\
    \hline
    \end{tabular}    
  \end{table}
  For comparison, two i-vector baselines are shown.
  The first baseline is based on~\cite{Dehak:2011}, and
  uses 13 PLPs with first-order and second-order derivatives,
  1024 Gaussians, and 300-dimensional i-vectors. 
  The second baseline is based on~\cite{Garcia:2011} with
   150 eigenvoices. 
   The i-vector+PLDA baseline should be taken with a grain of salt as the PLDA model was only trained on a subset of the 2M\_train data set (4k speakers and 50 utterances per speaker) due to limitations of our current implementation.\footnote{However, training with only 30 utterances per speaker gives almost the same results.}
  Also, this baseline does not include other refining techniques such as 
  "uncertainty training"~\cite{Stafylakis:2013} that
  have been reported to give substantial additional gains under certain conditions.
  Note that compared to~\cite{Variani:2014}, we have improved our d-vectors 
  significantly~\cite{Chen:2015}.

%\vspace{-1ex}	
\subsection{Softmax vs. End-to-End Loss}\label{sec:losses}
%\vspace{-1ex}
Next, we compare the softmax loss (Section~\ref{sec:speaker-verification}) and 
end-to-end loss (Section~\ref{sec:e2e}) for training utterance-level speaker representations.
Table~\ref{tab:clean-eval} shows the equal error rates for the DNN in Fig.~\ref{fig:dnn}.
If trained on the small training set (train\_2M), 
the error rates on the raw scores are comparable for
the different loss functions. 
While dropout gives a 1\% absolute gain for softmax,
we did not observe a gain from dropout for the end-to-end loss.
Similarly, t-normalization helps by 20\% for softmax, but not at all for the end-to-end loss.
This result is in agreement with the degree of consistency between the training 
loss and the evaluation metric.
In particular, the end-to-end approach assuming a global threshold in training 
(see Eq.~(\ref{eq:e2e-loss})), can implicitly learn normalized scores that are invariant under
different noise conditions etc. and makes score normalization redundant.
When using the softmax DNN for initialization of the end-to-end training, 
the error rate is reduced from 2.86\% to 2.25\%, suggesting an estimation problem.

If trained on the larger training set (train\_22M), 
the end-to-end loss clearly outperforms softmax, see Table~\ref{tab:clean-eval}.
To reasonably scale the softmax layer to 80k speaker labels, we employed 
candidate sampling, similar to~\cite{Jean:2014}.
Again, t-normalization helps by 20\% for softmax and softmax can catch up with
the other losses, which do not benefit from t-normalization. 
The initialization for end-to-end training (random vs. "pre-trained" softmax DNN)
does not make a difference in this case.

Although the step time for the end-to-end approach is larger than for softmax
with candidate sampling because the speaker model is computed on the fly, 
the overall convergence times are comparable.

\begin{table}[h]\vspace{-2ex}
    \centering
    \caption{\label{tab:clean-eval} Equal error rates for different losses,
    $^{\star}$ is with candidate sampling.}
    \begin{tabular}{|l|r|r|r|}
      \hline
                 & \multicolumn{2}{c|}{EER (\%), raw / t-norm} \\
      loss       & train\_2M & train\_22M \\
      \hline\hline
      softmax    & 2.90 /  2.28    & 2.69 / 2.08$^{\star}$ \\
      end-to-end & 2.86 / 2.85     & 2.04 / 2.14\phantom{$^{\star}$} \\
      \hline
    \end{tabular}    
  \end{table}

\begin{figure}
\centering
\includegraphics[width=0.4\textwidth]{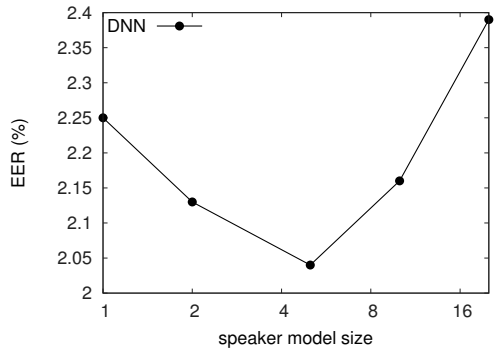}
\vspace{-2ex}
\caption{\it Speaker model size vs. equal error rate (EER).}
\label{fig:model-size}
\vspace{0ex}
\end{figure}

 The optimal choice of the number of utterances used to estimate the speaker model
 in training,
 referred to as the speaker model size, depends on the (average) number of enrollment utterances.
% see Fig.~\ref{fig:enroll-utts-per-speaker} for a histogram on our clean enrollment data.
%\begin{figure}
%\centering
%\includegraphics[width=0.5\textwidth]{figures/enrollment-utterances-histogram}
%\caption{\it Number of speakers for a given number of enrollment utterances,  
%the average speaker model size is 6.}
%\label{fig:enroll-utts-per-speaker}
%\end{figure} 
 In practice, however, smaller speaker model sizes may be more attractive to reduce
 the training time and make the training harder. 
Fig.~\ref{fig:model-size} shows the dependency of the test equal error rate on
the speaker model size, i.e., the number of utterances used to estimate the speaker model.
There is a relatively broad optimum around a model size of 5
with 2.04\% equal error rate, compared to 2.25\% for a model size of 1.
This model size is close to the true average model size, which is 6 for our enrollment set.
Similar trends are observed for the other configurations in this paper (not shown).
This indicates the consistency of the proposed training algorithm with the
verification protocol and suggests that task-specific training tends to be better.

%\vspace{-1ex}
\subsection{Feedforward vs. Recurrent Neural Networks}\label{sec:architectures}
%\vspace{-1ex}
So far we focused on the "small footprint" DNN in Fig.~\ref{fig:dnn} with
one locally-connected and three fully-connected hidden layers.
Next, we explore larger and different network architectures, regardless of their size and
computational complexity. The results are summarized in Table~\ref{tab:architectures}.
Compared to the small footprint DNN, the "best" DNN uses an additional hidden layer
and gives a 10\% relative gain.
%A convolutional neural network (CNN) obtained from the DNN by replacing the
%locally-connected layer with a convolutional layer and a pooling layer, including
%subtractive normalization, is worse than the "small footprint" DNN in terms
%of error rate, computational complexity, and convergence rate, 
%see Section~\ref{sec:speaker-verification} for further comments.
The LSTM in Fig.~\ref{fig:lstm} adds another 30\% gain over this best DNN.
The number of parameters is comparable to that of the DNN but the LSTM involves about
ten times more multiplications and additions. More hyperparameter tuning
will hopefully bring the computational complexity further down to make it feasible.
Slightly worse error rates are achieved with the softmax loss
(using t-normalization, candidate sampling, dropout, and possibly early stopping, which
were all not needed for the end-to-end approach).
On train\_2M, we observed similar relative gains in error rate over the 
respective DNN baselines.

\begin{table}[h]\vspace{-2ex}
    \centering
    \caption{\label{tab:architectures} Equal error rates
    for different model architectures using end-to-end training, 
    $^\dagger$ is with t-norm score normalization and trained only on the smaller training set.}
    \begin{tabular}{|l|r|}
      \hline
                 & EER (\%) \\
      \hline\hline
      frame-level DNN baseline & 3.32$^\dagger$ \\
      \hline
      DNN, "small footprint"   & 2.04\phantom{$^\dagger$} \\
      DNN, "best"                  & 1.87\phantom{$^\dagger$} \\
      LSTM                           & 1.36\phantom{$^\dagger$} \\
      \hline
    \end{tabular}    
  \end{table}

  \section{Summary \& Conclusion}\label{sec:summary}
  We proposed a novel end-to-end approach to speaker verification, which
  directly maps the utterance to a score and
  jointly optimizes the internal speaker representation and the speaker model
  using the same loss for training and evaluation.
  Assuming sufficient training data, the proposed approach 
  improved our best small footprint DNN baseline from over 3\% to 2\% equal error rate
  on our internal "Ok Google" benchmark.
  Most of the gain came from the utterance-level vs. frame-level modeling.
  Compared to other losses, the end-to-end loss achieved the same or
  slightly better results but with fewer additional concepts.
  In case of softmax, for example, we obtained comparable error rates only when using 
  score normalization at runtime, candidate sampling to make training feasible, 
  and dropout in training.
  Furthermore, we showed that the equal error rate can further be reduced to 1.4\%
  using a recurrent neural network instead of a simple deep neural network, although
  at higher computational runtime cost.
  By comparison, a reasonable but not fully state-of-the-art i-vector/PLDA system gave 4.7\%.
  Clearly, more comparative studies are needed.
  Nevertheless, we believe that our approach demonstrates a promising
  new direction for big data verification applications.

  %% \section{Acknowledgements}
  
  %%   The ISCA Board would like to thank the organizing committees of the past INTERSPEECH conferences for their help and for kindly providing the template files.

 % \newpage
  \ninept
  \bibliographystyle{IEEEtran}

  \bibliography{paper}

\end{document}